\documentclass[10pt,twocolumn,letterpaper,table]{article}

\usepackage{cvpr}              % To produce the CAMERA-READY version
\definecolor{cvprblue}{rgb}{0.21,0.49,0.74}
\usepackage[pagebackref,breaklinks,colorlinks,allcolors=cvprblue]{hyperref}
\usepackage{xcolor}
\usepackage{booktabs}
\usepackage{multirow}

\def\confName{CVPR}
\def\confYear{2025}

\title{\textit{ConceptGuard}: Continual Personalized Text-to-Image Generation\\ with Forgetting and Confusion Mitigation}

\author{\textbf{Zirun Guo}\quad \textbf{Tao Jin}\\
Zhejiang University\\
{\tt\small zrguo.cs@gmail.com}
}

\begin{document}
\maketitle

\begin{abstract}
Diffusion customization methods have achieved impressive results with only a minimal number of user-provided images. However, existing approaches customize concepts collectively, whereas real-world applications often require sequential concept integration. This sequential nature can lead to catastrophic forgetting, where previously learned concepts are lost. In this paper, we investigate concept forgetting and concept confusion in the continual customization. To tackle these challenges, we present \textbf{ConceptGuard}, a comprehensive approach that combines shift embedding, concept-binding prompts and memory preservation regularization, supplemented by a priority queue which can adaptively update the importance and occurrence order of different concepts. These strategies can dynamically update, unbind and learn the relationship of the previous concepts, thus alleviating concept forgetting and confusion.
Through comprehensive experiments, we show that our approach outperforms all the baseline methods consistently and significantly in both quantitative and qualitative analyses.
\end{abstract}

\section{Introduction}
Text-to-Image (T2I) diffusion models~\citep{nichol2022glide, rombach2022high, saharia2022photorealistic, podellsdxl} have emerged as a promising approach in the field of generative artificial intelligence, enabling the creation of high-quality images from textual descriptions. These models find a wide range of applications across various domains. Among their many uses, customization stands out as one of the most practical and popular applications. By harnessing the capabilities of T2I diffusion models, diffusion customization allows for the generation of user-defined concepts, achieving impressive results~\citep{ruiz2023dreambooth, galimage, kumari2023multi, li2024photomaker, wei2023elite, linnon}.

\begin{figure}
  \centering
  \includegraphics[width=0.8\linewidth]{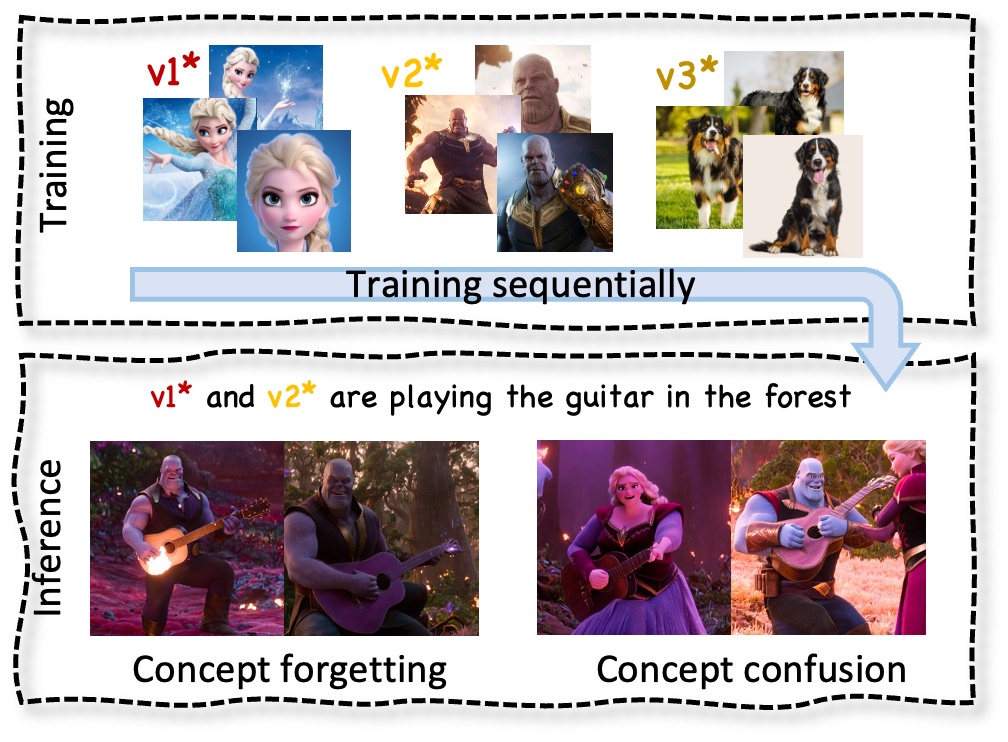}
  \caption{Examples of catastrophic forgetting and concept confusion in existing methods under the continual training setting.}
  \label{intro}
  \vskip -0.15 in
\end{figure}

However, in real-world applications, the new concepts we aim to customize typically arrive in a sequence. Existing methods~\citep{ruiz2023dreambooth, galimage, kumari2023multi, wei2023elite, linnon} primarily focus on how to personalize these concepts collectively. This will give rise to problems when they are applied to a dynamic and continual environment. For example, as shown in Figure~\ref{intro}, when generating previous concepts, these methods might experience catastrophic forgetting~\citep{mccloskey1989catastrophic} or concept confusion. Concretely, catastrophic forgetting occurs when the model fails to retain previously learned concepts, resulting in an inability to generate them accurately. Concept confusion arises when the model struggles to differentiate between earlier concepts, leading to outputs that blend elements of different concepts. In such continual settings, when fine-tuned on new concepts, the update of model parameters will influence the generation of previous concepts, thus leading to these issues. 

To address the forgetting problem in continual diffusion customization, Continual Diffusion~\citep{Smith2023ContinualDC} proposes C-LoRA, composed of a continually self-regularized low-rank adaptation in cross attention layers of the diffusion models. However, Continual Diffusion only explores how to deal with the forgetting problem, overlooking the concept confusion in the continual environment. As shown in Figure~\ref{multi}, combinations of multiple concepts are still generated wrongly or blended together. Based on these observations, we propose \textbf{\textit{ConceptGuard}}, consisting of several simple yet effective strategies to address these issues. Specifically, we propose shift embeddings, concept-binding prompts and memory preservation regularization along with a priority queue which connects all the strategies. Shift embedding helps to dynamically update the concept embedding in the continual environment. Concept-binding prompt strategy consists of concept focus, chrono-concept composition and concept-binding prompts, which helps to unbind the concepts and assess the importance and relationships of different concepts. Memory preservation regularization prevents the model from updating too fast which leads to catastrophic forgetting. Furthermore, we employ a priority queue capable of adaptively assessing the importance and occurrence order of different concepts to facilitate concept replay.

We conduct extensive experiments to validate the effectiveness of our method. Experimental results indicate that in a continual environment, whether for single-concept image generation or multi-concept image generation, our proposed method consistently outperforms existing approaches in both quantitative and qualitative analyses, producing images of high quality. Ablation experiments are then conducted to validate the effectiveness of different strategies of our method.
Our contributions can be summarized as:
\begin{itemize}
  \item In addition to catastrophic forgetting, we observe that concept confusion in continual customization causes elements to blend, a limitation that existing methods have yet to overcome.
  \item To mitigate concept forgetting and confusion, we propose a comprehensive approach that combines shift embedding, concept-binding prompts, and memory preservation regularization, supplemented by a priority queue.
  \item Our method consistently outperforms existing approaches in both quantitative and qualitative analyses, producing images of high quality.
\end{itemize}

\section{Related Work}
\noindent\textbf{Text-to-Image Diffusion Models.}
Diffusion models~\citep{nichol2022glide, rombach2022high, podellsdxl} have demonstrated remarkable abilities in generation tasks. Text-to-Image (T2I) diffusion models aims to generate images from textual descriptions generated by a pre-trained text encoder. GLIDE~\citep{nichol2022glide} adopts CLIP guidance and classifier-free guidance for text-conditioned image synthesis. Imagen~\citep{saharia2022photorealistic} leverages pre-trained large language models to provide rich textual information for image generation. Latent diffusion models such as Stable Diffusion~\citep{rombach2022high} propose to project images into latent representations and add text as conditional information to the denoising process. Stable Diffusion XL~\citep{podellsdxl} further expands the parameters of the model and achieves better results.

\noindent\textbf{Customization in Diffusion Models.} Diffusion customization~\citep{ruiz2023dreambooth, galimage, kumari2023multi, li2024photomaker, wei2023elite, linnon, lin2024action} aims to generate user-defined concepts under various contexts. Textual Inversion~\citep{galimage} adds tokens of new concepts to dictionary and fine-tunes these new tokens. DreamBooth~\citep{ruiz2023dreambooth} proposes a class-specific prior preservation loss to safeguard prior knowledge and conducts fine-tuning of all parameters of Stable Diffusion. Custom Diffusion~\citep{kumari2023multi} proposes to fine-tune the cross attention layers and combine multiple fine-tuned models into one via closed-form constrained optimization. However, in real-world continual applications where new concepts come continually, fine-tuning the model to new concepts would give rise to forgetting of previous customized concepts.

\noindent\textbf{Continual Learning.} Continual Learning methods can be roughly divided into three categories. (1) Regularization methods~\citep{kirkpatrick2017overcoming, li2017learning, zenke2017continual} address catastrophic forgetting by imposing a regularization constraint to important parameters. (2) Replay-based methods~\citep{rebuffi2017icarl, chaudhryefficient, bang2021rainbow} store some representative samples of previous tasks in a memory buffer and retrain these samples. (3) Architecture approaches~\citep{yoon2018lifelong, von2020continual} dynamically expand the network to mitigate forgetting for different tasks. Recently, prompt learning~\citep{wang2022learning, yan2024low, guo2025efficient} emerges as a kind of efficient methods to address catastrophic forgetting.

\noindent\textbf{Continual Customization.} Continual customization~\citep{Smith2023ContinualDC, 10489849, dong2024continually} aims to personalize concepts in a sequential manner. \citet{Smith2023ContinualDC} propose a regularization term to avoid forgetting in diffusion models. \citet{10489849} address forgetting in diffusion models by adding a task-aware memory buffer. \citet{dong2024continually} introduce concept consolidation loss and an elastic weight aggregation module to address concept forgetting and neglect. However, these methods always encounter concept confusion, failing in some complex compositions of multiple concepts.

\section{\textit{ConceptGuard}}

\begin{figure*}
    \centering
    \includegraphics[width=0.85\linewidth]{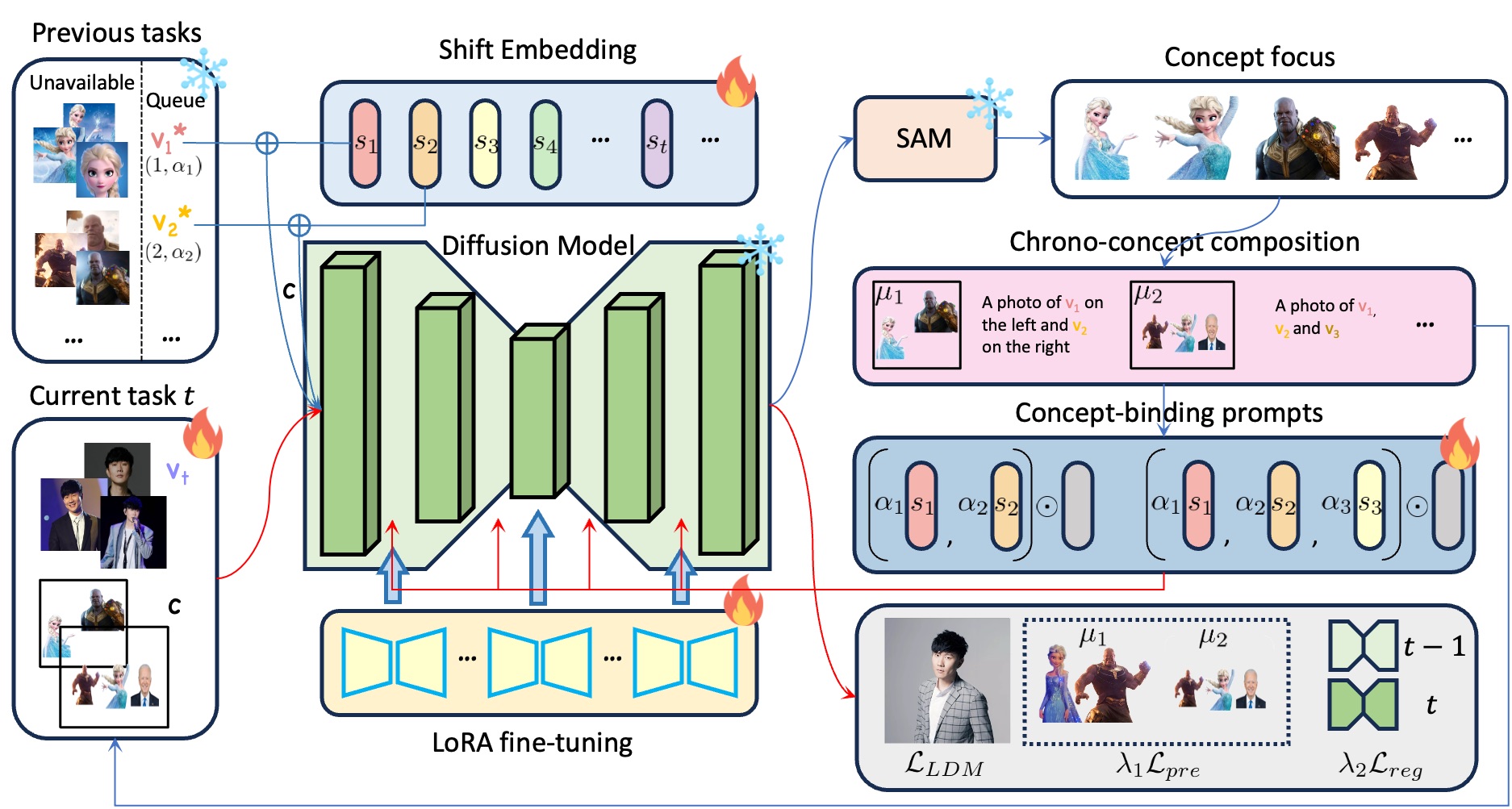}
    \caption{Overall architecture of our proposed method ConceptGuard.}
    \label{overall}
    \vskip -0.15 in
\end{figure*}

\subsection{Background}\label{s31}

\noindent\textbf{Latent Diffusion Models.} 
We implement our method on the publicly available Stable Diffusion XL model~\citep{podellsdxl}, which is a Latent Diffusion Model (LDM) for image-to-text synthesis. Specifically, LDM first projects the image $\mathbf{x}$ into a latent space feature $\mathbf{z} = \mathcal E (\mathbf{x})$ via an encoder $\mathcal E$. Then LDM performs the diffusion process in the latent space which can be divided into the forward process and the reverse process. The forward process adds Gaussian noise to the sample $\mathbf{z}$ to obtain $\mathbf{z}_t$: $q(\mathbf{z}_t|\mathbf{z})=\mathcal N (\mathbf{z}_t; \sqrt{\bar{\alpha}_t}\mathbf{z}_0, (1-\bar{\alpha}_t)\mathbf I)$, where $\mathbf{z_t}$ is the noised sample at time step $t$, $\bar{\alpha}_t=\Pi_{i=1}^t \alpha_i$ and $\alpha_i$ is the noise schedule parameter~\citep{ho2020denoising}. The reverse process iteratively removes the added noise in the forward process to obtain $\mathbf{z}_0$:  $p_\theta (\mathbf{z}_{t-1}|\mathbf{z}_t)=\mathcal N (\mathbf{z}_{t-1}; \mu_\theta (\mathbf{z}_t, t), \sigma_t)$, where $\mu_\theta (\mathbf{z}_t, t)= \frac{1}{\sqrt{\alpha_t}}(\mathbf{z}_t - \frac{1-\alpha_t}{\sqrt{1-\bar{\alpha}_t}}\epsilon_\theta(\mathbf{z}_t,t))$, $\sigma_t = \frac{1-\alpha_{t-1}}{1-\bar{\alpha}_t}\beta_t$, $\beta_t=1-\alpha_t$ and $\epsilon_\theta$ is the network used to predict the added noise. Then, LDM is trained with the loss function:
\begin{equation}\label{e1}
    \mathcal{L}_{\text{LDM}}=\mathbb{E}_{\mathbf{z}\sim\mathcal{E}(\mathbf{x}),c,\varepsilon\sim\mathcal{N}(0,1),t}\left[\left\|\varepsilon-\epsilon_\theta\left(\mathbf{z}_t,t,c\right)\right\|_2^2\right]
\end{equation}
where $t$ is the timestep and $(\mathbf{z}_t, c)$ are the corresponding pairs of image latents and text embeddings.

\noindent\textbf{Problem Formulation.} Given a series of concepts $\{\mathcal D^k \}_{k=1}^K$ where each concept $\mathcal D^k$ consists of 3-5 images $x_i^k$ with text prompts $c_i^k$, continual customization aims to personalize these concepts sequentially without forgetting previous concepts. Considering the privacy or storage issues, previous images are unavailable at current task.

\subsection{Overview}\label{s32}
We focus on the two challenges in Figure~\ref{intro}. The first is catastrophic forgetting, which occurs when the model fails to retain previously learned concepts. The second is concept confusion, where the model struggles to differentiate the current concept from similar past concepts, resulting in the generation of blended outputs. Textual Inversion~\cite{galimage} only trains the added tokens without fine-tuning the diffusion models which can avoid the forgetting to some extent. However, compared to methods fine-tuning the diffusion models, keeping the model fixed is not an ideal solution because it can not capture the characteristics of the concepts accurately. Besides, as shown in Figure~\ref{multi}, it fails in multi-concept generation. Therefore, following previous method~\citep{kumari2023multi}, we fine-tune the $K$ and $V$ matrices of self-attention layers in the diffusion model with LoRA~\citep{hulora}.

To address catastrophic forgetting and concept confusion during the process of model fine-tuning, we propose shift embedding, concept-binding prompts and memory preservation regularization, supplemented by a priority queue. Figure~\ref{overall} presents the overall architecture of ConceptGuard.

\subsection{Shift Embedding}\label{s33}
\begin{figure}
    \centering
    \includegraphics[width=0.72\linewidth]{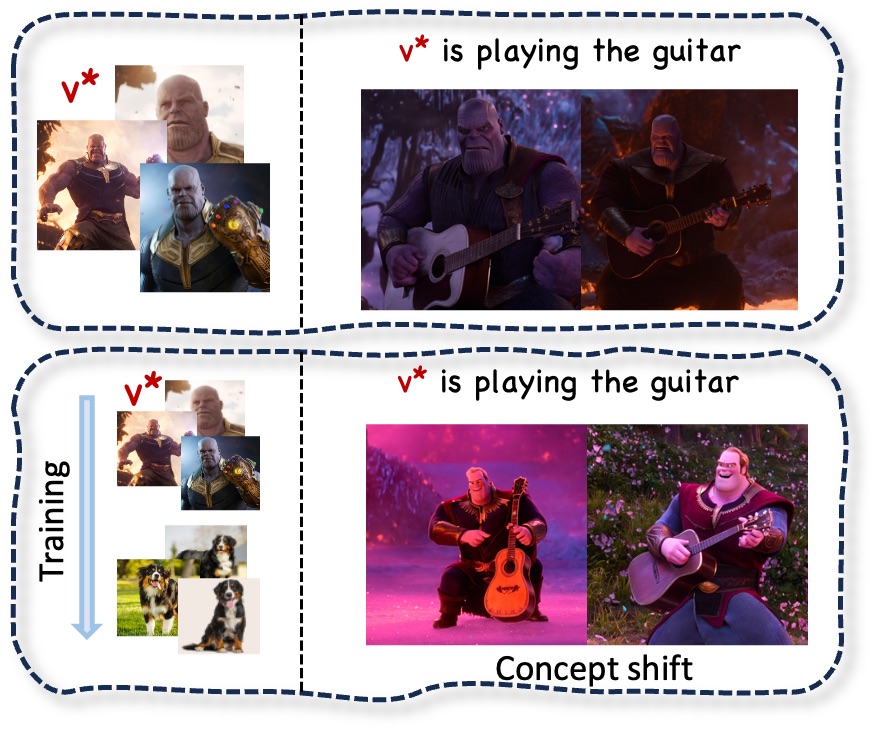}
    \caption{Examples of concept shift under the continual settings.}
    \label{cshift}
    \vskip -0.15 in
  \end{figure}

Textual Inversion~\citep{galimage} introduces a new token and the corresponding text embedding for each concept. During customization, it keeps the model fixed and only trains the text embedding of the new concept. Following Textual Inversion, we also introduce a new token and the corresponding text embedding for each concept. Additionally, to capture the characteristics of the new concept accurately, we also fine-tune the diffusion model following~\citep{kumari2023multi}. However, as shown in Figure~\ref{cshift}, in a continual environment, the fine-tuning of the model results in the previously learned text embeddings not accurately representing the features of the concepts. We call this phenomenon in continual diffusion as \textit{concept shift}. Based on this observation, we propose shift embeddings to fit the update of the model adaptively.

Suppose the current task is $\mathcal D^t$, then we have a newly added corresponding text embedding $v_t$ and the previous fine-tuned concept embeddings $v^*_1, v^*_2, \cdots, v^*_{t-1}$. For each concept $v_i$, we introduce a dynamic trainable weight term $\alpha_i$ to measure the importance of $v_i$ among all the customized concepts. For each previous concept, we introduce a tuple $(i, \alpha_i)$ where $i$ denotes the order of the concept and $\alpha_i$ represents its importance. We establish a priority queue to store all the concept tuples. When customizing the current concept, we extract 3 to 5 previous concepts from the queue. The criteria for extraction prioritize the order of the concepts based on $i$; concepts with smaller $i$ values are placed at the front of the queue. In cases where $i$ values are equal, concepts with higher importance $\alpha_i$ are positioned ahead in the queue. After extracting these concepts from the queue, we set $i$ to the current concept index $t$ and dynamically update $\alpha_i$ (see Section~\ref{s34}).

Subsequently, we create prompts for the selected concepts, such as \texttt{"a photo of a $v^\prime_i$"} and input them into the diffusion model to generate the corresponding images. Here, the text embedding $v^\prime_i$ is calculated as:
\begin{equation}
    v^\prime_i = v^*_i + s_i
\end{equation}
where $v^*_i$ is the fine-tuned text embeddings from previous concept $i$, $s_i$ is the shift embedding which is initialized to a zero vector and $i<t$. During the customization of the current concept $t$, we only fine-tune the shift embeddings $s_i,i=1,2,\cdots,t-1$ of selected concepts and the text embedding $v_t$ of current concept. By employing this approach, we enable the text embeddings of previously learned concepts to evolve in a continual environment alongside the model's updates, thereby preventing catastrophic forgetting.

\subsection{Concept-binding Prompts}\label{s34}
Shift embedding takes the update of the model into account and shift the text embeddings accordingly, thus addressing the catastrophic forgetting problem. However, as we observed in Figure~\ref{intro}, concept fusion between concepts is also a problem. During the customization process, the model can not distinguish the importance of each concept adaptively. Specifically, customization makes the model overfit to some specific concepts, especially simple and distinctive concepts. Therefore, we propose three simple yet effective strategies to deal with the concept confusion problem.

\noindent\textbf{Concept focus.} From the generated images, we can observe that the model will generate similar backgrounds to that of the training data if no prompt related to background information is added. For example, when the prompt \texttt{"a photo of $v_1$"} is given to generate an image of $v_1$, the background is always depicted as a snowy landscape. This has a negative impact on the model, as it becomes unclear what should be remembered and what should be forgotten in a continual environment. Therefore, in the continual environment, we want the model to focus on the main concept in images. We leverage a pre-trained model SAM~\citep{kirillov2023segment} to remove the background and keep the concept.

\noindent\textbf{Chrono-concept composition.} As introduced in Section~\ref{s33}, we establish a priority queue to select previous concepts according to their order and importance. After concept focus, we randomly combine these selected concepts (2-5 concepts) to create new images with different contexts. For each concept, we set a fixed weight $\mu$ based on the order in which the concept appears. Specifically, concepts that are further away from the current concept receive the highest weight, while those that are closer to the current concept receive the lowest weight. We set the weight of the nearest concept to 0.2 and the weight of the farthest concept to 0.6. The weights of the intermediate concepts are determined through linear interpolation based on their order. Then, for each image, we sum the weights of the concepts to obtain the overall weight of that image.

\noindent\textbf{Concept-binding prompts.} One of the main reasons for concept confusion is that, with the arrival of new concepts, the model struggles to assess the importance of different concepts and to clarify the relationships between them. Motivated by prompt learning~\citep{voynov2023p+, guo2024multimodal, yan2024low}, we propose concept-binding prompts to address the problem. Specifically, we first introduce a trainable weight term $\alpha$ for each concept which is initialized to 1. This weight term enables the model to distinguish the importance of each concept dynamically in the continual environment. For an image from chrono-concept composition which consists of a series of concepts $\mathcal C$, we calculate its concept-binding prompts $P$ as:
\begin{equation}\label{e3}
    P = [\alpha_c \cdot s_c]_{c\in\mathcal C}\odot P_b
\end{equation}
where $s_c$ is the shift embedding of concept $c$, $[\cdot]_{c\in\mathcal C}$ is the concatenation operation for all concepts in $\mathcal C$, $\odot$ is the broadcasting element-wise multiplication and $P_b$ denotes the trainable global binding prompts for all concepts. Concretely, $P\in\mathbb R^{\ell\times d}$ and $P_b\in\mathbb R^{d}$ where $\ell$ is the number of concepts in the image and $d$ is the dimension of the word embedding. Then, the concept-binding prompt $P$ will be inputted into the diffusion model as conditions. By introducing concept-binding prompts, we can not only dynamically update the importance of different concepts in the continual environment but also can connect different concepts through the global binding prompts $P_b$, which helps reduce the catastrophic forgetting and concept confusion. Meanwhile, the dynamic weight term $\alpha$ will be used in the priority queue in Section~\ref{s33}.

\subsection{Memory Preservation Regularization}\label{s35}
Following previous personalization methods~\citep{kumari2023multi,linnon,Smith2023ContinualDC}, we also use LoRA~\citep{hulora} for model fine-tuning. LoRA decomposes the original weight into two low-rank matrices: $\mathbf W^\prime = \mathbf W + \Delta \mathbf W$ where $\Delta \mathbf W = \mathbf A  \mathbf B^\top$. LoRA only fine-tunes the low-rank matrices $\mathbf A$ and  $\mathbf B$ and thus reduces the number of parameters. To further prevent the overfitting of current concept and prevent the model from updating too fast which leads to catastrophic forgetting, we propose memory preservation regularization for LoRA matrices which can be formalized as:
\begin{equation}
    \mathcal L_{reg} = \frac{1}{L}\sum_{\ell=1}^L \Vert \Delta\mathbf W_\ell^{t-1} - \Delta\mathbf W_\ell^{t} \Vert^2
\end{equation}
where $\ell$ denotes the LoRA layer and $t$ is the task index. By imposing the memory preservation regularization, we slow down the update of the diffusion models, which is crucial to maintain the knowledge of previous concepts.

\subsection{Training}\label{s36}
We take the customization of the $t$-th concept as an example. We first select several previous concepts from the priority queue. Then, we sum the word embeddings of these concepts with their corresponding shift embeddings and input them into the diffusion model to generate the corresponding images. Afterward, we utilize the SAM~\citep{kirillov2023segment} for concept focus and perform chrono-concept composition to obtain each image along with the associated prompt and weight. Based on the combinations of different concepts in each image, we derive the concept-binding prompt using Equation~\ref{e3}. We pair these images with their corresponding prompts $c$, weights $\mu$, and concept-binding prompts $P$, along with the current concept $t$, for the customization training of the current concept. For the current concept $t$, we use $\mathcal L_{LDM}$ in Equation~\ref{e1} for optimization. For previous concepts, we denote the loss function as $\mathcal L_{pre}$ which additionally adds concept-binding prompts as the conditions and image weight $\mu$ to $\mathcal L_{LDM}$ for different images. Besides, the memory preservation regularization term is added to form the overall loss function:
\begin{equation}\label{losseq}
    \mathcal L = \mathcal L_{LDM} + \lambda_1 \mathcal L_{pre} + \lambda_2 \mathcal L_{reg}
\end{equation}
where $\lambda_1$ and $\lambda_2$ are the loss trade-offs. During the customization of concept $t$, we only fine-tune the LoRA layers, shift embeddings $s$ and weight $\alpha$ of previous concepts, text embedding $v_t$ of current concept and the global binding prompts $P_b$. After the training of each concept, the weight $\alpha$ will be updated and used to select concepts from the priority queue in the next customization round. Meanwhile, $(t, \alpha_t)$ will be added into the queue where $\alpha_t$ is initialized to 1. When inference, we add shift embedding $s$ to the concept embedding $v^*$ as the final embedding.

\section{Experiments}
\subsection{Experimental Settings}
\noindent\textbf{Datasets.} We follow the settings of previous work~\citep{kumari2023multi,ruiz2023dreambooth} and select 18 concepts for customization, which contains person, pets, cartoon characters, etc. For evaluation, we select 6 concepts for continual customization each time.

\noindent\textbf{Evaluation metrics.} To evaluate the quality of the generated images, we follow previous work~\citep{ruiz2023dreambooth,kumari2023multi} and adopt \textit{Image-alignment} (IA, the visual similarity of generated images with the target concept, using similarity in CLIP~\citep{Radford2021LearningTV} image feature space) and \textit{Text-alignment} (TA, using text-image similarity in CLIP feature space). Besides, to evaluate the ability to address catastrophic forgetting, we adopt forgetting measure. Specifically, we introduce \textit{Forgetting-Image} (FI) and \textit{Forgetting-Text} (FT) which can be computed as $\text{FI}_j^t=\max_{i\in \{1,\cdots,t-1\}} (\text{IA}_{i,j}-\text{IA}_{t,j})$ and $\text{FT}_j^t=\max_{i\in \{1,\cdots,t-1\}} (\text{TA}_{i,j}-\text{TA}_{t,j})$ where $\text{IA}_{i,j}$, $\text{TA}_{i,j}$ denote the IA and TA of the $j$-th concept after customizing the $i$-th concept and $\text{IA}_{t,j}$, $\text{TA}_{t,j}$ denote the IA and TA of the $j$-th concept after customizing the $t$-th concept. We calculate the average of $\text{FI}_j^t$ and $\text{FT}_j^t$ of all concepts after customizing all the concepts as FI and FT.

\begin{table}[]
    \centering
    \caption{Quantitative comparisons between different methods.}
    \label{comp}
    \resizebox{0.9\columnwidth}{!}{%
    \begin{tabular}{@{}l|cc|cc|cc@{}}
    \toprule
    \multirow{2}{*}{Method}& \multicolumn{2}{c}{\textit{Text Alignment}} & \multicolumn{2}{c}{\textit{Image Alignment}} & \multirow{2}{*}{FT($\downarrow$)} & \multirow{2}{*}{FI($\downarrow$)} \\ \cmidrule(r){2-5}
     & single & multi & single & multi &  &  \\ \midrule
    Textual Inversion & 40.1 & 35.1 & 71.1 & 45.3 & \textbf{0.0} & \textbf{0.0} \\
    DreamBooth & 41.5 & 36.2 & 73.2 & 49.8 & 2.7 & 6.8 \\
    - with EWC & 41.8 & 36.7 & 75.3 & 54.4 & 2.4 & 6.1 \\
    - with LwF & 42.7 & 37.1 & 77.7 & 56.7 & 2.1 & 5.2 \\
    Custom Diffusion& 42.5 & 36.4 & 72.4 & 50.1 & 2.5 & 6.4 \\
    - with EWC & 41.5 & 36.1 & 74.9 & 55.1 & 2.2 & 5.5 \\
    - with LwF & 42.6 & 37.2 & 75.8 & 56.8 & 1.9 & 4.6 \\
    Continual Diffusion & 42.3 & 37.8 & 77.5 & 57.1 & 1.7 & 4.1 \\ \midrule
    \rowcolor{green!40}\textbf{\textit{ConceptGuard}} & \textbf{43.1} & \textbf{40.3} & \textbf{81.3} & \textbf{69.8} & 0.9 & 1.9 \\ \bottomrule
    \end{tabular}%
    }
    \vskip -0.15 in
\end{table}

\begin{figure*}
    \centering
    \includegraphics[width=0.85\linewidth]{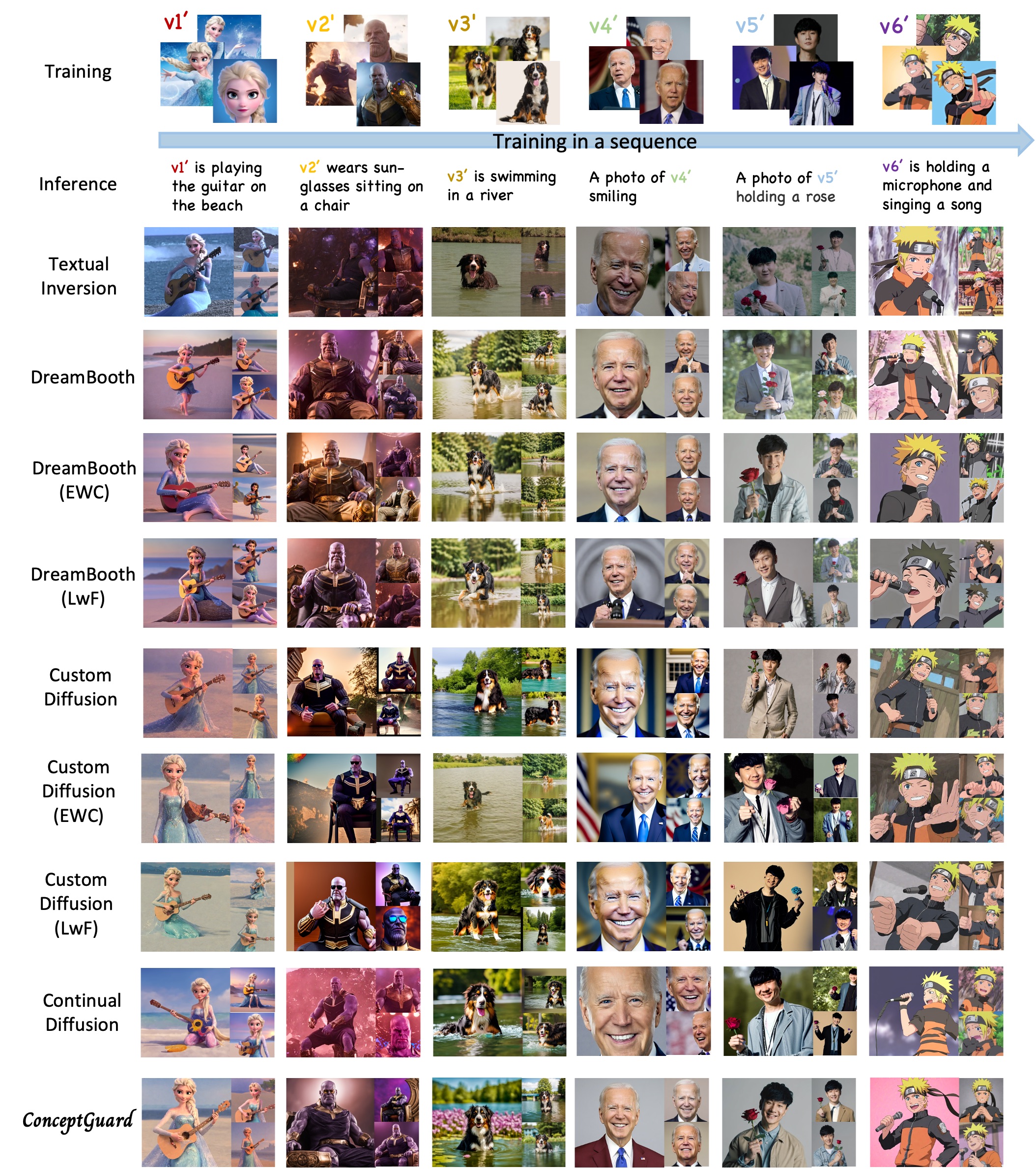}
    \caption{Comparison of single-concept generation of different methods.}
    \label{single}
    \vskip -0.15 in
\end{figure*}

\begin{figure*}
    \centering
    \includegraphics[width=0.85\linewidth]{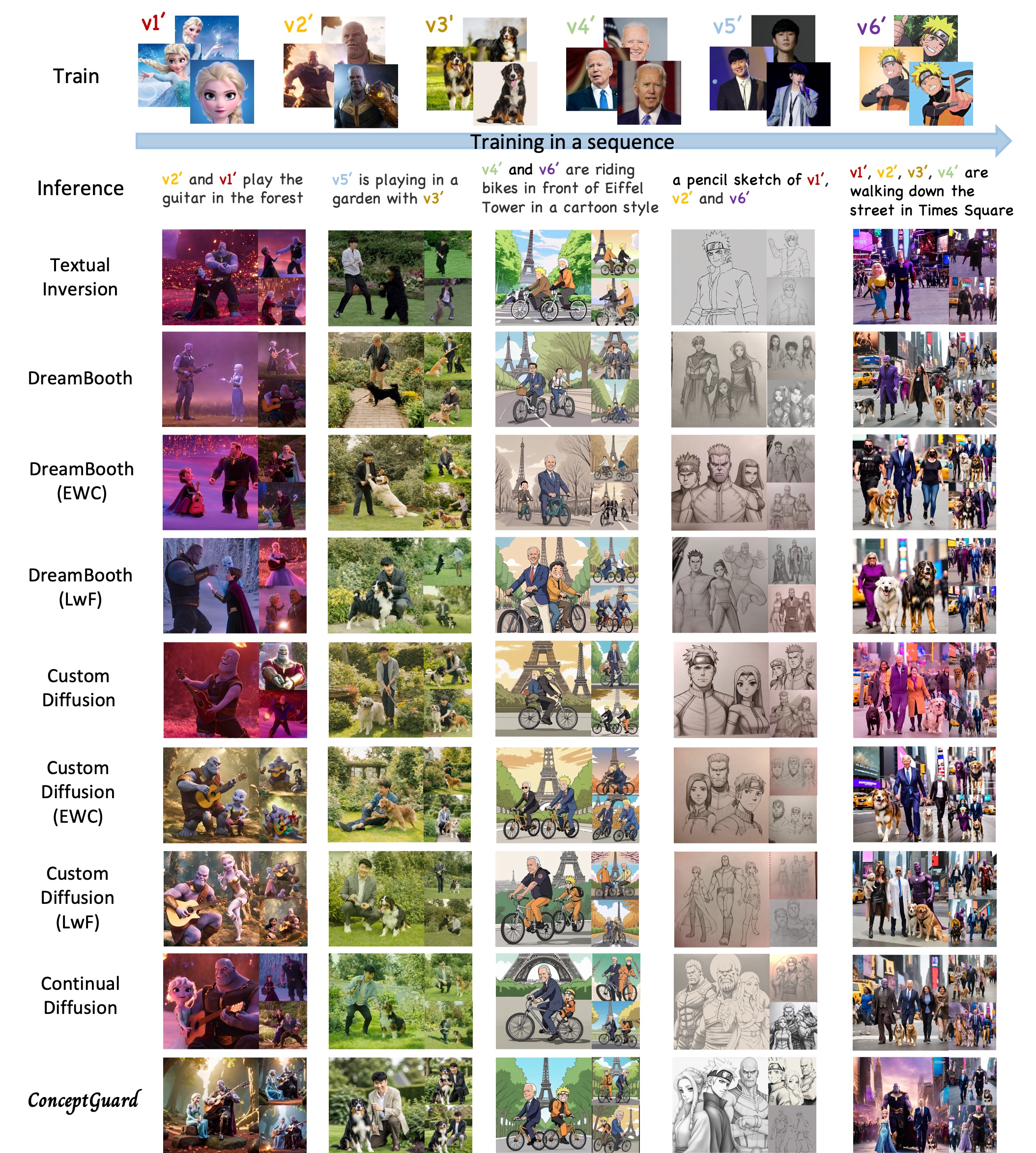}
    \caption{Comparison of multi-concept generation of different methods.}
    \label{multi}
    \vskip -0.15 in
\end{figure*}

\noindent\textbf{Implementation details.} We compare our method with four different baselines, Textual Inversion~\citep{galimage}, DreamBooth~\citep{ruiz2023dreambooth}, Custom Diffusion~\citep{kumari2023multi} and Continual Diffusion~\citep{Smith2023ContinualDC}. For non-continual methods Textual Inversion, DreamBooth and Custom Diffusion, we apply two different continual learning strategies, EWC~\citep{kirkpatrick2017overcoming} and LwF~\citep{li2017learning}. We use SDXL~\citep{podellsdxl} as the pre-trained diffusion model. For each task, we fine-tune the model for 600 steps with a learning rate of $1\times 10^{-4}$ and batch size of one. The weight terms $\lambda_1$ and $\lambda_2$ of the loss function are set to 1 and 0.5 by default.

\subsection{Qualitative Comparison}
We present our single-concept and multi-concept results in Figure~\ref{single} and \ref{multi}, respectively. From Figure~\ref{single}, we can observe that our method demonstrates better performance in generating images that not only adhere more closely to the textual descriptions but also exhibit a higher level of realism. Besides, methods with continual learning strategies struggle to capture the details of concepts. From Figure~\ref{multi}, we can observe that existing methods exhibit catastrophic forgetting and concept fusion. This results in outputs where elements from multiple concepts are blended together, creating a lack of clarity and coherence. Additionally, these methods frequently struggle to generate certain concepts entirely, leading to incomplete or inaccurate representations. In contrast, our method accurately captures the importance and relationships between different concepts, enabling it to generate images that closely align with the provided text descriptions. This is achieved through our innovative strategies, including shift embedding to dynamically adjust the concept embedding, concept-binding prompts which interact between different concepts and adjust the importance of different concepts, and the memory preservation regularization to preserve the learned knowledge.

\subsection{Quantitative Comparison}
Following \citep{kumari2023multi}, we use 20 text prompts and 50 samples per prompt for each concept for single-concept evaluation and 10 prompts containing complex interactions between concepts for multi-concept evaluation. In Table~\ref{comp}, we present the alignment scores and forgetting scores of different methods. From the results, we have the following observations: (1) existing methods exhibit significant shortcomings in multi-concept generation during continual training.  Their performance drops considerably compared to single-concept generation; (2) the forgetting metrics of Textual Inversion are 0 because Textual Inversion only fine-tunes the added tokens while keeping the diffusion model frozen. However, this will yield lower alignment scores and image quality because the added token can not capture the characteristics of the concepts accurately; (3) our method excels in both single- and multi-concept generation, demonstrating strong resistance to forgetting. Moreover, our approach incorporates shift embedding, which effectively mitigates the negative impact of regularization on the model's performance in accurately generating the details of concepts. 

\begin{table}[]
    \centering
    \caption{Effectiveness of different components of our method.}
    \label{abex}
    \resizebox{0.9\columnwidth}{!}{%
    \begin{tabular}{@{}l|cc|cc|cc@{}}
    \toprule
    \multirow{2}{*}{Method}& \multicolumn{2}{c}{\textit{Text Alignment}} & \multicolumn{2}{c}{\textit{Image Alignment}} & \multirow{2}{*}{FT($\downarrow$)} & \multirow{2}{*}{FI($\downarrow$)} \\ \cmidrule(r){2-5}
     & single & multi & single & multi &  &  \\ \midrule
    w/o. SE & 42.4 & 38.0 & 78.9 & 63.4 & 1.7 & 3.1 \\
    w/o. PQ & 42.8 & 39.7 & 80.4 & 67.1 & 1.1 & 2.4 \\
    w/o. CBP & 42.1 & 37.8 & 78.1 & 59.3 & 2.4 & 3.9 \\
    w/o. MPR & 42.7 & 38.9 & 79.7 & 66.5 & 1.2 & 2.7 \\\midrule
    \rowcolor{green!40}\textbf{\textit{ConceptGuard}} & \bf{43.1} & \bf{40.3} & \bf{81.3} & \bf{69.8} & \bf{0.9} & \bf{1.9} \\ \bottomrule
    \end{tabular}%
    }
    \vskip -0.15 in
\end{table}

\subsection{Ablation Study}
We conduct several ablation experiments of shift embedding (SE), priority queue (PQ), concept-binding prompts (CBP) and memory preservation regularization (MPR).

\begin{figure}
    \centering
    \includegraphics[width=0.92\linewidth]{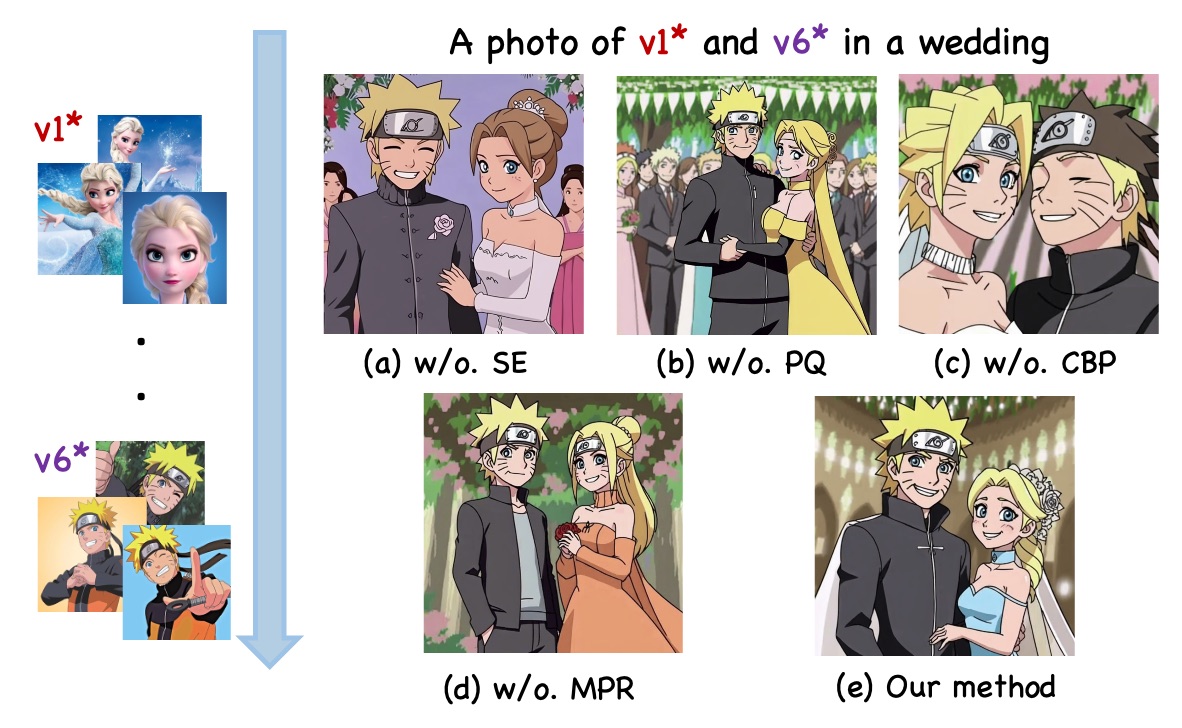}
    \vskip -0.1 in
    \caption{Ablation experiments of our method.}
    \label{ab1fig}
    \vskip -0.05 in
\end{figure}

\begin{figure}
    \centering
    \includegraphics[width=0.55\linewidth]{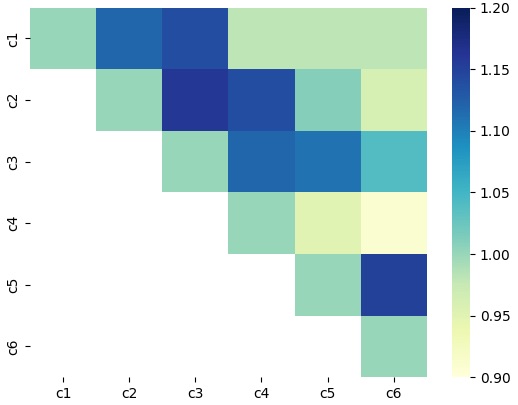}
    \caption{Concept importance during the continual customization. The value of $i$-th row and $j$-th column denotes the importance of concept $i$ after customizing concept $j$.}
    \label{ab4fig}
    \vskip -0.05 in
\end{figure}

\begin{figure}
    \centering
    \includegraphics[width=0.93\linewidth]{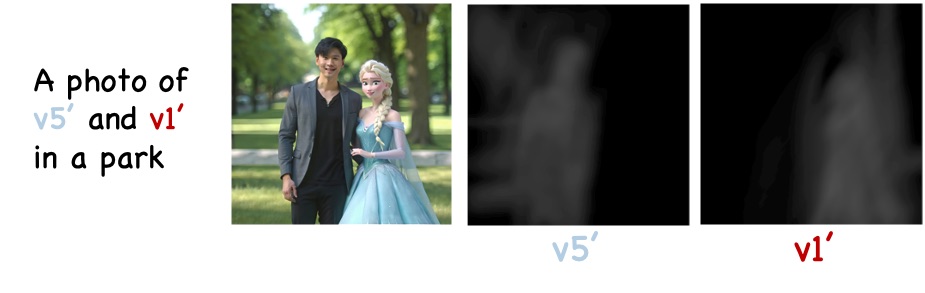}
    \vskip -0.1 in
    \caption{Visualizations of concept embeddings.}
    \label{attnmap}
    \vskip -0.15 in
\end{figure}

\noindent\textbf{Effectiveness of different components.} In Table~\ref{abex} and Figure~\ref{ab1fig}, we present the quantitative and qualitative results of different components in ConceptGuard. We can observe that concept-binding prompts bring the most improvements to the model. This is because concept-binding prompt strategy includes concept focus, chrono-concept compositions which can unbind the concepts, thus alleviating the concept confusion. We also visualize the concept importance $\alpha$ during the customization in Figure~\ref{ab4fig}. We can observe that the model can adaptively adjust the importance of concepts where important concepts will be prioritized. Besides, we visualize the cross-attention map of concept embeddings in Figure~\ref{attnmap}. This figure indicates the effectiveness of our concept-binding prompts and shift embeddings.
Furthermore, all the strategies help to alleviate the concept forgetting and improve the quality of generated images, demonstrating the effectiveness of our proposed strategies.

\begin{figure}
    \centering
    \includegraphics[width=\linewidth]{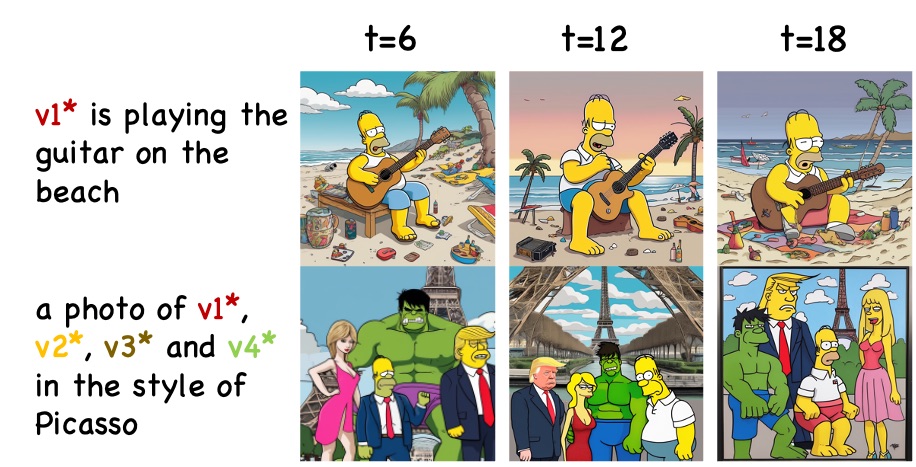}
    \caption{Visualizations after customizing the concept $t$.}
    \label{ab2fig1}
    \vskip -0.1 in
\end{figure}

\begin{figure}
    \centering
    \includegraphics[width=\linewidth]{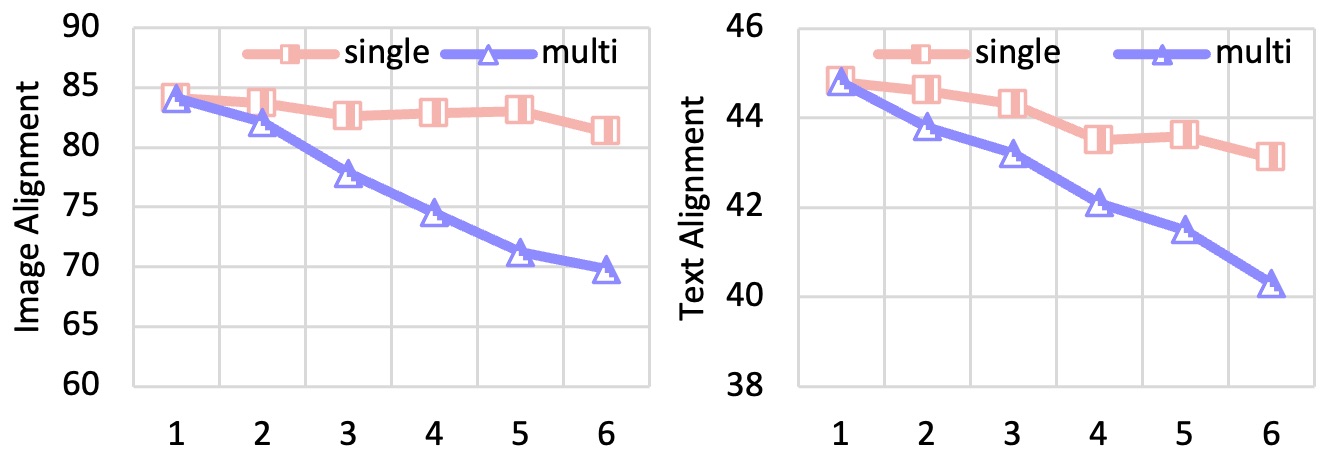}
    \caption{Performance of different numbers of concepts.}
    \label{ab2fig}
    \vskip -0.1 in
\end{figure}

\noindent\textbf{Number of learned concepts.} In Figure~\ref{ab2fig}, we present the changes in performance given different numbers of learned concepts. Besides, in Figure~\ref{ab2fig1}, we present the results of more concepts. We can observe that with the increase of concepts, our method can remain stable performance, demonstrating its effectiveness and robustness.

\begin{figure}
    \centering
    \includegraphics[width=\linewidth]{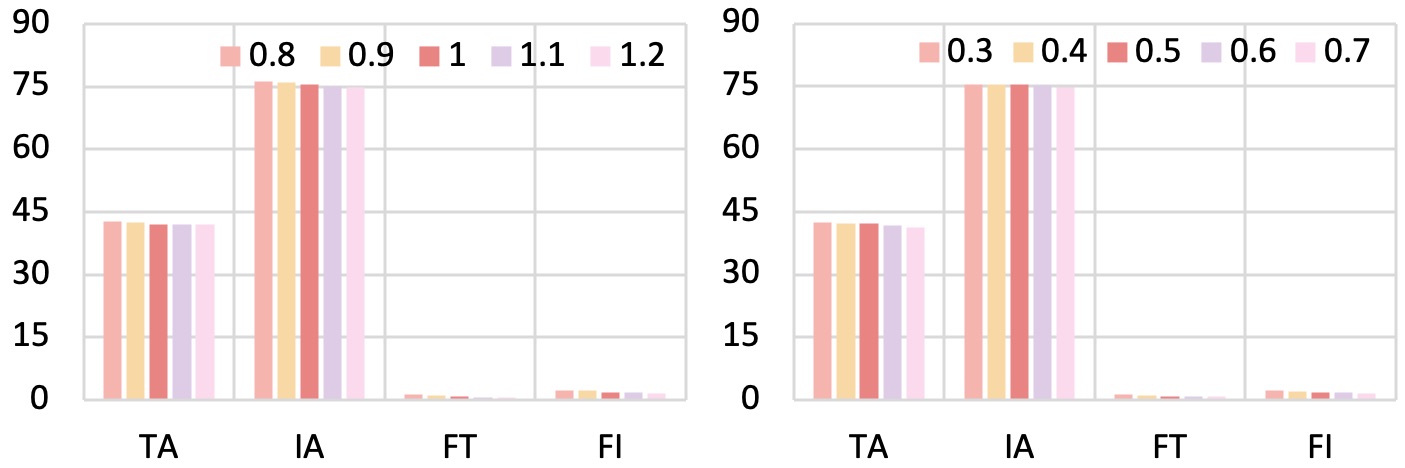}
    \caption{Ablation experiments of $\lambda_1$ (left) and $\lambda_2$ (right).}
    \label{ab3fig}
    \vskip -0.15 in
\end{figure}

\noindent\textbf{Loss trade-off.} We select several values of $\lambda_1$ and $\lambda_2$ in Equation~\ref{losseq} and present the results in Figure~\ref{ab3fig}. With the increase of $\lambda_1$ and $\lambda_2$, the alignment scores decrease and the forgetting improve. This indicates that increasing $\lambda_1$ and $\lambda_2$ helps to alleviate the forgetting but sacrifice the quality of the generated images, because preventing the model from updating fast might lead to underfitting of current concepts. However,  our method can achieve stable performance given different values $\lambda$, demonstrating its robustness.

\section{Conclusion}
We investigate the concept forgetting and concept confusion problems under the context of continual customization. To tackle these challenges, we propose a comprehensive framework that combines shift embedding, concept-binding prompts and memory preservation regularization, supplemented by a priority queue which can adaptively update the importance and occurrence order of different concepts. Extensive experiments demonstrate that our method consistently and significantly outperforms other methods in both quantitative and qualitative analyses.

{
    \small
    \bibliographystyle{ieeenat_fullname}
    \bibliography{main}
}


\end{document}